\documentclass[11pt]{article}

\usepackage[letterpaper,textwidth=6.1in,textheight=8.6in,centering]{geometry}
\usepackage{amsmath,amsfonts,amssymb,amsthm,amscd}
\usepackage{graphicx}
\usepackage{color}
\usepackage{mathrsfs}
\usepackage{bm}
\usepackage{enumerate}
\usepackage{algorithm}
\usepackage{algorithmic}
\usepackage{cite}

\numberwithin{equation}{section}
\newcommand{\bx}{\mathbf{x}}
\newcommand{\by}{\mathbf{y}}
\newcommand{\bn}{\mathbf{n}}

\newcommand{\p}{\partial}

\newcommand{\e}{\epsilon}

\newcommand{\mathd}{\mathrm{d}}

\DeclareMathOperator{\sgn}{sgn}
\DeclareMathOperator{\dist}{dist}

\theoremstyle{definition}

\title{A New Formulation of the Elastica Model with Application to Image Inpainting}
\author{Zuoqiang Shi\thanks{Yau Mathematical Sciences Center, Tsinghua University, Beijing 100084, China; Yanqi Lake Beijing Institute of Mathematical Sciences and Applications, Beijing 101408, China. Email: \texttt{zqshi@tsinghua.edu.cn}.}
\and Xue-Cheng Tai\thanks{NORCE Norwegian Research Centre, Nyg\aa rdsgaten 112, 5008 Bergen, Norway. Email: \texttt{xtai@norceresearch.no}.}}
\date{}

\begin{document}
\maketitle

\begin{abstract}
Curves and interfaces play a central role in computer vision and image processing, and Euler's elastica energy is one of the most successful tools to model and
manipulate them. The elastica energy is usually computed through a level-set or a phase field representation of the curve, in which the curvature term makes the
energy nonlinear, non-convex and, in the level-set case, non-differentiable, so that its minimization is notoriously difficult.
In this paper, we propose a new formulation of the elastica model based on a binary representation of curves. The key ingredient is a very simple formula for the
curvature: if $\phi$ is the $\pm1$ binary function associated with a curve $\Gamma$, then along $\Gamma$ the curvature is given by
$-\frac{1}{C_0\epsilon}\left(R_\epsilon *\phi\right)+O(\epsilon)$, where $R_\epsilon$ is a radial kernel supported in a ball of radius $\epsilon$ and $C_0$ is an explicit
constant. No derivative of $\phi$ is involved and the convolution is evaluated by FFT.
The resulting energy is a weighted total variation, so that formally the new formulation is a generalization of the well-known ROF model; a phase field type model is
obtained from the same curvature formula. Combining FFT with the split Bregman iteration, which for the present problem coincides with an augmented Lagrangian
method (ALM), we obtain efficient algorithms whose subproblems are either linear systems solved by conjugate gradient or explicit pointwise formulas.
Numerical experiments on the inpainting of binary images are presented to demonstrate the effectiveness of the proposed algorithms.
\end{abstract}


\section{Introduction}
For a curve or interface $\Gamma$ in $\mathbb{R}^2$, the elastica energy is defined as
\begin{align}
  \label{def:elastica}
  E(\Gamma)=\int_\Gamma (a+b\kappa(s)^2)\mathd s.
\end{align}
Here, $\mathd s$ is the arc length of $\Gamma$, $a$ and $b$ are two positive constants and
$\kappa(s)$ is the curvature of $\Gamma$ at $s$. The elastica energy is also easy to extend to a surface in $\mathbb{R}^3$, in which case $\kappa$ becomes the mean curvature of the surface.

In the field of image processing, Euler's elastica energy was first introduced by Nitzberg, Mumford, and
Shiota for segmenting an image into objects with different depths in the scene \cite{NMS93}. Since then, it has been
adapted to many fundamental problems in mathematical imaging. This includes image inpainting \cite{Chan02,Masnou02,Tai11},
image restoration \cite{Tai11,Ambrosio09,Ambrosio03,Bae11}, image zooming \cite{Tai11} and image segmentation \cite{Zhu07,Zhu13,Bae17}.

The level-set method provides a way to compute the elastica energy \cite{Osher87}. Let $\phi$ be a level-set function with $\Gamma=\{\phi=0\}$. The length of $\Gamma$ is
given by
$$|\Gamma|=\int_\Omega \delta(\phi(\bx))\,|\nabla \phi(\bx)|\mathd\bx,$$
while at every point where $\nabla\phi\ne 0$ the curvature of the level curve of $\phi$ is the divergence of the unit normal,
$$\kappa=\nabla\cdot \left(\frac{\nabla \phi}{|\nabla \phi|}\right),$$
with the orientation for which $\nabla\phi/|\nabla\phi|$ is the outward normal of $\{\phi<0\}$. This expression is closely related to the length: the first variation of the
total variation $\int_\Omega|\nabla\phi|\mathd\bx$ is exactly $-\kappa$. The elastica energy then becomes
\begin{align}
  \label{eq:elastica-level}
  E(\phi)=\int_\Omega \left(a+b\left|\nabla\cdot \left(\frac{\nabla \phi}{|\nabla \phi|}\right)\right|^2\right)\delta(\phi(\bx))\,|\nabla\phi(\bx)|\,\mathd \bx.
\end{align}
In practice, $\delta$ is replaced by a smooth approximation. If the factor $\delta(\phi)$ is omitted, \eqref{eq:elastica-level} becomes the sum of the elastica energies of
all level curves of $\phi$, which is the form used to inpaint grayscale images \cite{Chan02}.

The energy in \eqref{eq:elastica-level} is nonlinear and nonconvex. It even involves non-differentiable terms. It is not easy to minimize.
Recently, many methods have been developed to minimize curvature based functionals, including the multigrid
algorithm \cite{Brito10}, the homotopy method \cite{Yang12}, augmented Lagrangian method (ALM) based algorithms
\cite{Tai11,Wu10,Zhu13}, graph cut based algorithms \cite{Bae11,Elzehiry10} and convex relaxation approaches \cite{Bredies15,Strekalovskiy11,Schoenemann09,Schoenemann12}.

Fast minimization algorithms for Euler's elastica have also been developed, including a relaxed normal two-split method and a weighted total variation approach for inpainting \cite{YK16}, an operator-splitting method for image smoothing \cite{DGT19}, and a bilinear-decomposition method \cite{Liu24}.

Another way to approximate the curvature is given by the phase field model. In the phase field model, the length of the interface and the curvature are approximated by
$$|\Gamma|\sim\int_\Omega\left(\epsilon |\nabla\psi|^2+\frac{1}{\epsilon}W(\psi)\right)\mathd\bx,\quad \kappa\sim-\epsilon \Delta\psi+\frac{1}{\epsilon}W'(\psi),$$
both up to a normalization constant that depends on $W$. Here $\psi$ is a phase function associated to $\Gamma$, $\epsilon\ll 1$ is a positive parameter and $W(\psi)=(1-\psi^2)^2$ is a double-well
potential that forces the phase function $\psi$ towards the binary values $\pm1$. 
The phase function is smooth, so the curvature is again computed by differentiation, and the transition layer must be resolved by the grid, which forces $\e$ to stay
comparable to the mesh size.

In this paper, we take a different route. We keep the binary representation of the curve and show that the curvature along $\Gamma$ is, up to an explicit constant and an
$O(\e)$ error, a single convolution of the binary function with a compactly supported radial kernel of width $\e$. Three properties distinguish this formula from the
level-set and phase field ones:
\begin{enumerate}[(i)]
\item it involves no derivative of $\phi$, so no finite differences of a discontinuous function have to be formed; the only derivative left in the energy is the
$|\nabla\phi|$ coming from the co-area formula, which is exactly the total variation term;
\item it is linear in $\phi$, so the curvature term of the elastica energy is quadratic in $\phi$ and the energy takes the form of a total variation weighted by a
quadratic functional of $\phi$. Setting $b=0$ recovers the plain total variation, so that formally the new model is a generalization of the well-known ROF model, and the
algorithmic machinery developed for ROF type energies applies;
\item the convolution is diagonalized by the FFT, so each evaluation costs $O(N\log N)$ operations for an image with $N$ pixels, independently of $\e$.
\end{enumerate}
Based on this formula we propose two models: an ROF type model, in which the curvature weight multiplies $|\nabla\phi|$, and a phase field type model, in which it
multiplies $|\nabla\phi|^2$ and a double-well penalty keeps the interface sharp. Both are minimized by the split Bregman iteration \cite{Goldstein09}, which splits the
energy into subproblems that are either linear systems solved by conjugate gradient or explicit pointwise formulas. The effectiveness of the resulting algorithms is
demonstrated on the inpainting of binary images.

The price to pay is that the curvature formula is exact only on $\Gamma$: at a point at distance $d$ from the interface an additional term of size $d/\e^2$ appears, and in
the computation the binary constraint has to be relaxed, so that $\phi$ develops a thin transition layer instead of a jump. These limitations, and the resulting open
question of whether the proposed model converges to the elastica model as $\e\to 0$, are discussed in Section \ref{sec:limit}.

The rest of the paper is organized as follows. In Section \ref{sec:formulation}, the new formulation is presented: the nonlocal formula for the length, the curvature
formula \eqref{eq:curvature-intro} and the elastica energy \eqref{eq:energy-elastic-nonlocal-intro} that they lead to. The curvature formula is proved in
Section \ref{sec:curvature}. The ROF type model and its algorithm are given in Section \ref{sec:alg}, and the phase field type model in Section \ref{sec:l2}.
In Section \ref{sec:num}, the discretization and implementation details are discussed and the
numerical results are presented. At last, the limitations of the current model and some possible directions to improve it are discussed in Section \ref{sec:limit}.

\section{A New Formulation of the Elastica Energy}
\label{sec:formulation}
\begin{figure}[ht]
  \centering
  \begin{tabular}{c}
  \includegraphics[width=0.4\textwidth]{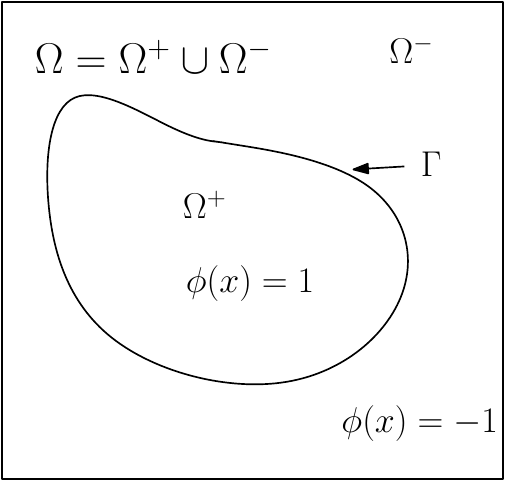}
   \end{tabular}
  \caption{Computational domain $\Omega=\Omega^+\cup\Gamma\cup\Omega^-$, the binary function $\phi$ and the curve $\Gamma$.}
  \label{fig:domain}
\end{figure}

Throughout the paper, $\Omega\subset\mathbb{R}^2$ is a bounded computational domain and $\Gamma\subset\Omega$ is a closed curve which splits $\Omega$ into an interior part
$\Omega^+$ and an exterior part $\Omega^-$, $\Omega=\Omega^+\cup\Gamma\cup\Omega^-$. Instead of a smooth level-set or phase function, we represent $\Gamma$ by the binary
function
\begin{equation}
  \label{eq:binary}
  \phi(\bx)=\left\{\begin{array}{rl}
  1,& \bx\in\Omega^+,\\
  -1,& \bx\in\Omega^-,
  \end{array}\right.
\end{equation}
so that $\Gamma$ is the jump set of $\phi$; see Fig. \ref{fig:domain}. Note that $\phi=\sgn(\phi)$ for such a function.

\paragraph{The kernel.} All the formulas below are built from one convolution kernel,
\begin{equation}
  \label{eq:kernel-intro}
  R_\e(\bx,\mathbf{y})=\frac{1}{\pi D_0 \e^2}R\left(\frac{|\bx-\by|^2}{\e^2}\right),\quad D_0=\int_0^1 R(s)\mathd s,
\end{equation}
where $R: \mathbb{R}^+\rightarrow \mathbb{R}^+$ is a positive function whose support is $[0,1)$. Thus $R_\e(\bx,\cdot)$ is supported in the ball $B_\e(\bx)$, and the
normalization in \eqref{eq:kernel-intro} is chosen so that $\int_{\mathbb{R}^2}R_\e(\bx,\by)\mathd\by=1$. Since $R_\e$ depends on $\bx-\by$ only, we also write
$R_\e(\bx-\by)$ for $R_\e(\bx,\by)$.
The simplest choice of $R$, and the one used in all our computations, is the Heaviside function
\begin{align}
  \label{eq:kernel-R}
  R(r)=\left\{\begin{array}{cc}
1,& 0\le r< 1,\\
0,& r\ge 1.
\end{array}
\right.
\end{align}
It is convenient to name the two convolution operators associated with $R_\e$,
\begin{equation}
  \label{eq:op-K}
  K_\e \phi(\bx)=\int_\Omega R_\epsilon(\bx-\by)  \phi(\by) \mathd \by,\qquad L_\e=\frac{1}{\e}K_\e.
\end{equation}
Since the kernel $R_\e$ is symmetric, both operators are self-adjoint, and both are evaluated by FFT.

\paragraph{A nonlocal formula for the length.} Our curvature formula comes from the variation of the length, so we first give a formula for the length of $\Gamma$ in terms
of $\phi$. It is easy to verify that
\begin{align}
  \label{eq:length-sign}
  \frac{1}{\epsilon}\int_\Omega\bigl(1+\sgn(\phi(\bx))\bigr)\left(\int_\Omega R_\e (\bx,\by)\bigl(1-\sgn(\phi(\by))\bigr)\mathd \by\right)\mathd\bx=4C_0|\Gamma|+O(\e),
\end{align}
where the constant $C_0>0$ is given in \eqref{eq:curv-c-intro} below.
Formula \eqref{eq:length-sign} is easy to understand: since $1+\sgn(\phi)=2$ in $\Omega^+$ and $1-\sgn(\phi)=2$ in $\Omega^-$, its left hand side equals
$$\frac{4}{\e}\int_{\Omega^+}\int_{\Omega^-}R_\e(\bx,\by)\,\mathd\by\,\mathd\bx,$$
i.e. it only counts pairs of points that are separated by $\Gamma$. Such a pair lies within
distance $\e$ of $\Gamma$, so the double integral scales like $\e$ times the length of $\Gamma$, which is what the factor $\frac{1}{\e}$ compensates.
Note that only two-point interactions enter \eqref{eq:length-sign}; no derivative of $\phi$ is involved.

\paragraph{From the length to the curvature.} Taking the variation of \eqref{eq:length-sign} with respect to $\phi$ and using $\sgn'=2\delta$, we get a formula for the
curvature
\begin{align}
  \label{eq:curv-intro}
  \frac{\delta |\Gamma|}{\delta \phi}(\bx)&\sim -\frac{1}{C_0\epsilon}\delta(\phi(\bx))\int_\Omega R_\e (\bx,\by)\phi(\by)\mathd \by.
\end{align}
This computation is only formal, since $\phi$ is a binary function; it is used as a motivation. Comparing with the classical level-set identity
$\frac{\delta |\Gamma|}{\delta \phi}\sim\delta(\phi)\kappa$, it suggests that the convolution above,
rescaled by $-1/(C_0\e)$, approximates the curvature on $\Gamma$. This is indeed the case: in Section \ref{sec:curvature}, we will prove that for $\bx\in \Gamma$
\begin{equation}
  \label{eq:curvature-intro}
  \frac{1}{\epsilon}\int_\Omega R_\e(\bx,\mathbf{y})\phi(\mathbf{y})\mathd \mathbf{y}=-C_0\kappa(\bx)+O(\e)
\end{equation}
with
\begin{equation}
  \label{eq:curv-c-intro}
 C_0=\frac{1}{\pi D_0}\int_0^1 \left(\int_{r^2}^1R(s)\mathd s \right)\mathd r=\frac{2}{\pi D_0}\int_0^1 r^2R(r^2)\mathd r>0.
\end{equation}
The constant in the length formula \eqref{eq:length-sign} is exactly $4C_0$ with the same $C_0$; this is what makes the formal variation \eqref{eq:curv-intro} consistent
with \eqref{eq:curvature-intro}. The sign in \eqref{eq:curvature-intro} depends on the orientation convention fixed in Section \ref{sec:curvature}; it is immaterial for
the models below, in which only $\kappa^2$ appears.

\paragraph{The elastica energy.} Formula \eqref{eq:curvature-intro} gives a very simple way to compute the curvature of the interface determined by a binary function
$\phi$: one convolution, no differentiation. Together with the co-area formula, which gives $\int_\Omega f|\nabla\phi|\mathd\bx=2\int_\Gamma f\mathd s$ for such a $\phi$,
it turns the elastica energy \eqref{def:elastica} into
\begin{equation}\label{eq:energy-elastic-nonlocal-intro}
E(\phi) = \int_\Omega \left(a+b \left | \frac{1}{\epsilon}\int_\Omega R_\epsilon(\bx-\by)  \phi(\by) \mathd \by  \right |^2\right) | \nabla  \phi(\bx) | \mathd \bx,
\end{equation}
where $\phi$ is a $\pm1$ binary function. The constants relating \eqref{eq:energy-elastic-nonlocal-intro} to \eqref{def:elastica}, namely the factor $C_0^{-2}$ from
\eqref{eq:curvature-intro} and the factor $\frac{1}{2}$ from the co-area formula, are absorbed into the parameters $a$ and $b$.
The energy \eqref{eq:energy-elastic-nonlocal-intro} is a total variation weighted by a quadratic functional of $\phi$; for $b=0$ it reduces to the total variation of
$\phi$, which is why we call it an ROF type model. It is studied, and applied to image inpainting, in Section \ref{sec:alg}. Written with the operators \eqref{eq:op-K}, the curvature term of
\eqref{eq:energy-elastic-nonlocal-intro} is $|L_\e\phi|^2|\nabla\phi|=\frac{1}{\e^2}|K_\e\phi|^2|\nabla\phi|$; replacing the factor $\frac{1}{\e^2}|\nabla\phi|$ by
$|\nabla\phi|^2$ and adding a double-well penalty to keep the interface sharp gives a second, phase field type model, which is studied in Section \ref{sec:l2}.
\section{Formula of Curvature}
\label{sec:curvature}
In this section, we derive formula \eqref{eq:curvature-intro} for the curvature of the interface $\Gamma$ from the binary function $\phi$ of \eqref{eq:binary}.

In addition to the setting of Section \ref{sec:formulation}, we assume from now on that $\Gamma\in C^2$.
We denote by $\bn$ the unit normal of $\Gamma$ pointing from $\Omega^+$ to $\Omega^-$ and by $\bm{\tau}$ the unit tangent, oriented so that the Frenet relations read
\begin{equation}
  \label{eq:frenet}
  \bm{\tau}'(s)=-\kappa(s)\bn(s),\qquad \bn'(s)=\kappa(s)\bm{\tau}(s).
\end{equation}
With this convention $\kappa>0$ when $\Omega^+$ is convex; for instance $\kappa\equiv 1/\rho$ if $\Omega^+$ is a disk of radius $\rho$.
We consider a point $\bx\in\Gamma$ with $\dist(\bx,\p\Omega)>\e$, so that the support of $R_\e(\bx,\cdot)$, which is the ball $B_\e(\bx)$, is contained in $\Omega$.

Let $\bar{R}(r)$ be the function defined by
\begin{equation}
  \label{eq:R-bar}
  \bar{R}(r)=-\frac{1}{2r}\int_r^{+\infty} R(s)\mathd s.
\end{equation}
Note that $\bar R\le 0$ and that, since the support of $R$ is $[0,1)$, the support of $\bar R$ is $[0,1)$ as well.
Direct calculation shows that, for any $\bx\ne 0$,
\begin{equation}
  \label{eq:divergence}
  \nabla\cdot\left(\frac{\bx}{\e} \,\bar{R}\left(\frac{|\bx|^2}{\e^2}\right)\right)=\frac{1}{\e}R\left(\frac{|\bx|^2}{\e^2}\right).
\end{equation}
We want to use this relation \eqref{eq:divergence} to do integration by parts to calculate the integral in \eqref{eq:curvature-intro}. Notice that $\bar{R}(r)$ is singular at
$r=0$. To do integration by parts, we have to exclude a small area around $\bx$. Denote $B_\sigma(\bx)$ as a small ball centered at $\bx$ with radius $\sigma$. It follows from the Gauss formula that
\begin{align}
  \label{eq:curv-1}
&  \pi D_0\e\int_{\Omega\backslash B_\sigma(\bx)} R_\e(\bx,\mathbf{y})\phi(\mathbf{y})\mathd \mathbf{y}\nonumber\\
=& -\int_{\Omega\backslash B_\sigma(\bx)} \phi(\by)\nabla_\by\cdot\left(\frac{\bx-\by}{\e} \,\bar{R}\left(\frac{|\bx-\by|^2}{\e^2}\right)\right)\mathd \by\nonumber\\
=& -2\int_{\Gamma_\sigma} \bn(\by)\cdot\left(\frac{\bx-\by}{\e} \,\bar{R}\left(\frac{|\bx-\by|^2}{\e^2}\right)\right)\mathd S_\by
-\int_{S^+_\sigma} \left(\frac{|\bx-\by|}{\e} \,\bar{R}\left(\frac{|\bx-\by|^2}{\e^2}\right)\right)\mathd S_\by\nonumber\\
&+\int_{S^-_\sigma} \left(\frac{|\bx-\by|}{\e} \,\bar{R}\left(\frac{|\bx-\by|^2}{\e^2}\right)\right)\mathd S_\by\nonumber\\
=& -2\int_{\Gamma_\sigma} \bn(\by)\cdot\left(\frac{\bx-\by}{\e} \,\bar{R}\left(\frac{|\bx-\by|^2}{\e^2}\right)\right)\mathd S_\by
-\frac{\sigma}{\e} \,\bar{R}\left(\frac{\sigma^2}{\e^2}\right)\left(\int_{S^+_\sigma} \mathd S_\by-\int_{S^-_\sigma} \mathd S_\by\right)
\end{align}
where $S^+_\sigma=\p B_\sigma(\bx)\cap \Omega^+$, $S^-_\sigma=\p B_\sigma(\bx)\cap \Omega^-$ and $\Gamma_\sigma=\Gamma\cap (\Omega\backslash B_\sigma(\bx))$.

From the assumption that $\Gamma\in C^2$, the two half circles $S^+_\sigma$ and $S^-_\sigma$ have lengths $\pi\sigma+O(\sigma^2)$, while
$\frac{\sigma}{\e}\bar R(\frac{\sigma^2}{\e^2})=O(\frac{\e}{\sigma})$ by \eqref{eq:R-bar}. Hence, as $\sigma\rightarrow 0$,
\begin{equation}
  \label{eq:integral-inner}
  \frac{\sigma}{\e} \,\bar{R}\left(\frac{\sigma^2}{\e^2}\right)\left(\int_{S^+_\sigma} \mathd S_\by-\int_{S^-_\sigma} \mathd S_\by\right)=O(\sigma)\rightarrow 0.
\end{equation}
Letting $\sigma$ go to $0$ in \eqref{eq:curv-1}, we obtain
\begin{equation}
\label{eq:curv-0}
  \frac{1}{\e}\int_{\Omega} R_\e(\bx,\mathbf{y})\phi(\mathbf{y})\mathd \mathbf{y}
  =-\frac{2}{\pi D_0\e^2}\int_{\Gamma} \bn(\by)\cdot\left(\frac{\bx-\by}{\e} \,\bar{R}\left(\frac{|\bx-\by|^2}{\e^2}\right)\right)\mathd S_\by.
\end{equation}
Let $s$ be the arclength of $\Gamma$ with $s=0$ at $\bx$, i.e. $\bx=\bx(0)$. To shorten the notation, we write
\begin{equation}
  \label{eq:R-short}
  \bar R_\e(s)=\bar{R}\left(\frac{|\bx(s)-\bx(0)|^2}{\e^2}\right).
\end{equation}
Recall that the support of $\bar R$ is $[0,1)$, so that only the part of $\Gamma$ inside
$B_\e(\bx)$ contributes. Using this fact, we have
\begin{align}
  \label{eq:int-arc}
  &-\int_{\Gamma} \bn(\by)\cdot\left(\frac{\bx-\by}{\e}\right) \,\bar{R}\left(\frac{|\bx-\by|^2}{\e^2}\right)\mathd S_\by\nonumber\\
=&\int_{\Gamma\cap B_\e(\bx(0))} \bn(s)\cdot\left(\frac{\bx(s)-\bx(0)}{\e} \right)\,\bar R_\e(s)\,\mathd s.
\end{align}
Next, we approximately compute the above integral by Taylor's expansion. From \eqref{eq:frenet},
\begin{equation}
  \label{eq:taylor-x1}
  \bx(s)-\bx(0) = s\bm{\tau}(0)-\frac{1}{2}\kappa(0)s^2\bn(0)+O(s^3),
\end{equation}
\begin{equation}
  \label{eq:taylor-n}
  \bn(s) = \bn(0)+\kappa(0)s\bm{\tau}(0)+O(s^2).
\end{equation}
Combining the above two expansions, we have
\begin{equation}
  \label{eq:taylor-x}
  \bn(s)\cdot(\bx(s)-\bx(0)) = \frac{1}{2}\kappa(0) s^2+O(s^3).
\end{equation}
Then, \eqref{eq:int-arc} gives that
\begin{align}
  \label{eq:curv-2}
  &\int_{\Gamma\cap B_\e(\bx(0))} \bn(s)\cdot\left(\frac{\bx(s)-\bx(0)}{\e}\right)\bar R_\e(s)\,\mathd s
=\frac{1}{2\e}\int_{\Gamma\cap B_\e(\bx(0))} \left(\kappa(0)s^2+O(s^3)\right) \bar R_\e(s)\,\mathd s
\end{align}
and, together with \eqref{eq:curv-0} and \eqref{eq:int-arc},
\begin{align}
  \label{eq:curv-3}
  \frac{1}{\e}\int_{\Omega} R_\e(\bx,\mathbf{y})\phi(\mathbf{y})\mathd \mathbf{y}
  =\frac{1}{\pi D_0\e^3}\int_{\Gamma\cap B_\e(\bx(0))} \left(\kappa(0)s^2+O(s^3)\right) \bar R_\e(s)\,\mathd s.
\end{align}
Next, we calculate the integrals $\int_{\Gamma\cap B_\e(\bx(0))} s^2 \bar R_\e(s)\mathd s$ and
$\int_{\Gamma\cap B_\e(\bx(0))} s^3 \bar R_\e(s)\mathd s$ respectively.

Let
$r(s)=|\bx(s)-\bx(0)|$. In $B_\e(\bx(0))$, we have
$$s=r+O(\e r),\quad \frac{\mathd s}{\mathd r}=1+O(\e).$$
Changing the integral variable to $r$, we get
\begin{align}
  \label{eq:integral-r-0}
\frac{1}{\e^3}\int_{\Gamma\cap B_\e(\bx(0))} s^2 \bar R_\e(s)\,\mathd s=
 (1+O(\e))\int_{-\epsilon}^{\e} \frac{r^2}{\e^3}  \bar{R}\left(\frac{r^2}{\e^2}\right)\mathd r.
\end{align}
Now, the integral is easy to compute,
\begin{align}
  \label{eq:integral-r-1}
  &\int_{-\epsilon}^{\e} \frac{r^2}{\e^3}  \bar{R}\left(\frac{r^2}{\e^2}\right)\mathd r=
  -2\int_{0}^{\e} \frac{r^2}{ \e^3}  \frac{\e^2}{2r^2}\left(\int_{r^2/\e^2}^1R(s)\mathd s \right)\mathd r=
  -\int_0^1 \left(\int_{r^2}^1R(s)\mathd s \right)\mathd r,
\end{align}
where the first equality comes from the definition of $\bar{R}$ and the second one from the substitution $r\rightarrow \e r$.

Now, we have
\begin{align}
  \label{eq:integral-r}
\frac{1}{\e^3}\int_{\Gamma\cap B_\e(\bx(0))} s^2 \bar R_\e(s)\,\mathd s=
-\int_0^1 \left(\int_{r^2}^1R(s)\mathd s \right)\mathd r+O(\e).
\end{align}
Given the above estimate, $\int_{\Gamma\cap B_\e(\bx(0))} s^3 \bar R_\e(s)\mathd s$ is easy to bound, since $|s|\le \e$ on
$\Gamma\cap B_\e(\bx(0))$,
\begin{align}
  \label{eq:integral-r-2}
  \left|\frac{1}{\e^3}\int_{\Gamma\cap B_\e(\bx(0))} s^3 \bar R_\e(s)\,\mathd s\right|\le
  \frac{1}{\e^2}\int_{\Gamma\cap B_\e(\bx(0))} s^2 \left|\bar R_\e(s)\right|\mathd s=O(\e).
\end{align}
Finally, \eqref{eq:curvature-intro} follows from \eqref{eq:curv-3}, \eqref{eq:integral-r} and \eqref{eq:integral-r-2}, with $C_0$ given by \eqref{eq:curv-c-intro}. The
second expression for $C_0$ in \eqref{eq:curv-c-intro} follows by exchanging the order of integration in the first one,
$$\int_0^1\left(\int_{r^2}^1R(s)\mathd s\right)\mathd r=\int_0^1R(s)\sqrt{s}\,\mathd s=2\int_0^1r^2R(r^2)\mathd r.$$

\section{ROF Type Model}
\label{sec:alg}
In this section we minimize the energy \eqref{eq:energy-elastic-nonlocal-intro} obtained in Section \ref{sec:formulation},
\begin{equation}\label{eq:energy-elastic-nonlocal}
E(\phi) = \int_\Omega \left(a+b \left | \frac{1}{\epsilon}\int_\Omega R_\epsilon(\bx-\by)  \phi(\by) \mathd \by  \right |^2\right) | \nabla  \phi(\bx) | \mathd \bx,
\end{equation}
and apply it to image inpainting.

In the image inpainting problem, the value of $\phi$ is given in a subset $\Omega_I\subset \Omega$. The purpose is to fill in the missing area to get a
complete image. In the elastica model, the missing part is filled by minimizing the elastica energy. More precisely, the image is reconstructed by solving the following
 optimization problem,
\begin{equation}\label{eq:opt-elastic-nonlocal}
\min_{\phi=\pm 1} \int_\Omega \left(a+b \left | \frac{1}{\epsilon}\int_\Omega R_\epsilon(\bx-\by)  \phi(\by) \mathd \by  \right |^2\right) | \nabla  \phi(\bx) | \mathd \bx
\end{equation}
with the constraint
\begin{equation}
  \label{eq:constraint}
  \phi(\bx)=\phi_0(\bx),\quad \bx\in \Omega_I.
\end{equation}
The binary constraint $\phi=\pm1$ is difficult to handle, so we relax it to the interval constraint $\phi\in [-1,1]$, which gives the following optimization problem,
\begin{equation}\label{eq:opt-elastic-nonlocal-relax}
\min_{\phi\in [-1,1]} \int_\Omega \left(a+b \left | \frac{1}{\epsilon}\int_\Omega R_\epsilon(\bx-\by)  \phi(\by) \mathd \by  \right |^2\right) | \nabla  \phi(\bx) | \mathd \bx
\end{equation}
with the constraint \eqref{eq:constraint}.

To simplify the notation, we use the operator $L_\e=\frac{1}{\e}K_\e$ of \eqref{eq:op-K}, which is self-adjoint, $L_\e^*=L_\e$.
We use the split Bregman method to solve the optimization problem \eqref{eq:opt-elastic-nonlocal-relax}. First, by introducing two auxiliary variables,
\eqref{eq:opt-elastic-nonlocal-relax} is rewritten as
\begin{align}\label{eq:opt-elastic-nonlocal-relax-bregman}
&\min_{\phi,\psi,D} \int_\Omega \left(a+b \left | L_\e(\phi)  \right |^2\right) | D(\bx) | \mathd \bx,\\
&\text{subject to:}\quad D(\bx)=\nabla  \phi(\bx),\quad \phi(\bx)=\phi_0(\bx),\quad \bx\in \Omega_I,\nonumber\\
& \quad\quad\quad\quad\quad\psi(\bx)=\phi(\bx),\quad -1\le\psi\le1.\nonumber
\end{align}
Using the split Bregman method, the above problem is split into three subproblems with respect to $\phi$, $\psi$ and $D$ respectively.
\begin{align}
  \label{eq:phi-bregman}
  \phi^{n+1}=&\arg\min_{\phi} \int_\Omega (a+b\left | L_\e(\phi)  \right |^2) | D^n(\bx)| \mathd\bx  +\frac{\alpha}{2}\int_\Omega |D^n(\bx)-\nabla \phi(\bx)+Q^n(\bx)|^2\mathd\bx\nonumber\\
    &\quad\quad +\frac{\beta}{2}\int_{\Omega} |\psi^n-\phi+P^n|^2\mathd\bx, \quad \mbox{subject to:}\quad \phi=\phi_0\; \mbox{in}\; \Omega_I.
\end{align}
Using a standard variational argument, $\phi^{n+1}$ is given by solving the linear equation
\begin{equation}
  \label{eq:phi}
  2b L_\e\left[|D^n(\bx)|L_\e \phi(\bx)\right]-\alpha\nabla\cdot\left[\nabla\phi(\bx)-D^n(\bx)-Q^n(\bx)\right]+\beta(\phi-\psi^n-P^n)=0,\quad \bx\in\Omega\backslash\Omega_I,
\end{equation}
together with the constraint
\begin{equation}
  \label{eq:boundary}
  \phi(\bx)=\phi_0(\bx),\quad \bx\in \Omega_I.
\end{equation}
This linear system is solved by conjugate gradient. Notice that $L_\e$ is a convolution operator which can be implemented efficiently by FFT, so that the above linear equation can be
solved very fast.

For $\psi$, we need to solve
\begin{align}
  \label{eq:psi-bregman}
  \psi^{n+1}=&\arg\min_{-1\le \psi\le 1} \frac{\beta}{2}\int_{\Omega} |\psi-\phi^{n+1}+P^n|^2\mathd\bx.
\end{align}
This problem has a closed form solution, the pointwise projection of $\phi^{n+1}-P^n$ onto $[-1,1]$,
\begin{align}
  \label{eq:psi-sol}
  \psi^{n+1}=\sgn(\phi^{n+1}-P^n)\min\{|\phi^{n+1}-P^n|,1\}.
\end{align}
To update $D$, an $\ell_1$ minimization problem has to be solved,
\begin{align}
  \label{eq:D-bregman}
  D^{n+1}=&\arg\min_{D} \int_\Omega (a+b\left | L_\e(\phi^{n+1})  \right |^2) | D(\bx)| \mathd\bx  +\frac{\alpha}{2}\int_\Omega |D(\bx)-\nabla \phi^{n+1}(\bx)+Q^n(\bx)|^2
    \mathd\bx.
\end{align}
It is well known that the solution of the above problem is given by soft thresholding,
\begin{align}
  \label{eq:D-sol}
  D^{n+1}(\bx)=S(\bx)\left(\nabla \phi^{n+1}(\bx)-Q^n(\bx)\right),
\end{align}
where the shrinkage factor $S(\bx)$ is given by
\begin{align}
  \label{eq:soft-shrink}
  S(\bx)=\max\left\{1-\frac{a+b\left | L_\e(\phi^{n+1})(\bx)\right|^2}{\alpha\left|\nabla \phi^{n+1}(\bx)-Q^n(\bx)\right|},0\right\},
\end{align}
with the convention $D^{n+1}(\bx)=0$ when $\nabla \phi^{n+1}(\bx)-Q^n(\bx)=0$.

\begin{algorithm*}[t]
\floatname{algorithm}{Algorithm}
\caption{Image Inpainting with Model \eqref{eq:opt-elastic-nonlocal-relax}}
\label{alg:inpainting}
\begin{algorithmic}[1]
\STATE Input: given data $\phi_0$ on $\Omega_I$, parameters $a, b, \alpha, \beta, \e$, initial values $D^0, \psi^0, Q^0, P^0$, error tolerance $\varepsilon$.
\REPEAT
\STATE Update $\phi^{n+1}$ by solving \eqref{eq:phi}.
\STATE Update $\psi^{n+1}$ by \eqref{eq:psi-sol}.
\STATE Update $D^{n+1}$ by \eqref{eq:D-sol}.
\STATE Update $Q^{n+1}$: $Q^{n+1}= Q^{n}+D^{n+1}-\nabla\phi^{n+1}.$
\STATE Update $P^{n+1}$: $P^{n+1}= P^{n}+\psi^{n+1}-\phi^{n+1}.$
\UNTIL $\|Q^{n+1}-Q^n\|+\|P^{n+1}-P^n\|<\varepsilon$.
\STATE Output: $\phi=\phi^{n+1}$.
\end{algorithmic}
\end{algorithm*}

\section{Phase Field Type Model}
\label{sec:l2}
Based on the curvature formula \eqref{eq:curvature-intro} derived in Section \ref{sec:curvature}, we can also get another type of model for the image inpainting problem.
In terms of the operator $K_\e=\e L_\e$ of \eqref{eq:op-K}, the model reads
\begin{equation}\label{eq:opt-elastic-l2}
\min_{\phi} \int_\Omega \left(\left |K_\e \phi(\bx)  \right |^2 | \nabla  \phi(\bx) |^2 +\frac{1}{\e}\left(1-\phi^2(\bx)\right)^2\right)\mathd \bx
\end{equation}
with the constraint
\begin{equation}
  \label{eq:constraint-l2}
  \phi(\bx)=\phi_0(\bx),\quad \bx\in \Omega_I.
\end{equation}
Comparing with the model \eqref{eq:opt-elastic-nonlocal-relax} of the previous section, the main difference is in the gradient term: the curvature term of
\eqref{eq:opt-elastic-nonlocal-relax} is $|L_\e\phi|^2|\nabla\phi|=\frac{1}{\e^2}|K_\e\phi|^2|\nabla\phi|$, whereas here it is $|K_\e\phi|^2|\nabla\phi|^2$, i.e. the large
factor $\frac{1}{\epsilon^2}$
is replaced by one extra power of $|\nabla \phi|$. In order to keep $\phi$ close to a function with a sharp interface, the penalty term $\frac{1}{\e}(1-\phi^2(\bx))^2$ is added to the energy.
By a simple scaling analysis, in this model $\phi$ should have a narrow layer of width $O(\e^2)$, which is consistent with the substitution
$\frac{1}{\e^2}\leftrightarrow|\nabla\phi|$ above.

This new model is easier to solve. We only need to introduce one auxiliary variable in the split Bregman iteration,
\begin{align}\label{eq:opt-elastic-l2-bregman}
&\min_{\phi,\psi} \int_\Omega \left(\left |K_\e \phi(\bx)  \right |^2 | \nabla\psi(\bx) |^2 +\frac{1}{\e}\left(1-\phi(\bx)\psi(\bx)\right)^2\right)
\mathd \bx,\\
&\text{subject to:}\quad \psi(\bx)=\phi(\bx),\quad \bx\in\Omega,\qquad \phi(\bx)=\phi_0(\bx),\quad \bx\in \Omega_I.\nonumber
\end{align}
Using the split Bregman method, the above problem is split into two subproblems with respect to $\phi$ and $\psi$ respectively.
The problem for $\phi$ is
\begin{align}
  \label{eq:phi-bregman-l2}
  \phi^{n+1}=&\arg\min_{\phi} \int_\Omega \left(\left |K_\e \phi(\bx)  \right |^2 | \nabla\psi^n(\bx) |^2
+\frac{1}{\e}\left(1-\phi(\bx)\psi^n(\bx)\right)^2\right)
\mathd \bx \nonumber\\
&\hspace{2cm}+\frac{\gamma}{2}\int_{\Omega} |\psi^n-\phi+P^n|^2\mathd\bx, \quad \mbox{subject to:}\quad \phi=\phi_0\; \mbox{in}\; \Omega_I.
\end{align}
Using a standard variational argument, $\phi^{n+1}$ is given by solving the linear equation
\begin{equation}
  \label{eq:phi-l2}
   2K_\e\left[|\nabla\psi^n(\bx)|^2 K_\e \phi(\bx)\right]-\frac{2}{\e}\psi^n(\bx)\left(1-\phi(\bx)\psi^n(\bx)\right)+\gamma(\phi-\psi^n-P^n)=0,\quad \bx\in\Omega\backslash\Omega_I,
\end{equation}
together with the constraint
\begin{equation}
  \label{eq:boundary-l2}
  \phi(\bx)=\phi_0(\bx),\quad \bx\in \Omega_I.
\end{equation}
As in Section \ref{sec:alg}, $K_\e$ is self-adjoint and can be evaluated by FFT, so \eqref{eq:phi-l2} can be solved efficiently by conjugate gradient.

For $\psi$, we need to solve
\begin{align}
  \label{eq:psi-bregman-l2}
  \psi^{n+1}=&\arg\min_{\psi} \int_\Omega \left(\left |K_\e (\phi^{n+1})  \right |^2 | \nabla\psi |^2 +\frac{1}{\e}\left(1-\phi^{n+1}(\bx)\psi(\bx)\right)^2\right)
\mathd \bx\nonumber\\
&\hspace{2cm}+\frac{\gamma}{2}\int_{\Omega} |\psi-\phi^{n+1}+P^n|^2\mathd\bx.
\end{align}
To get the solution, we need to solve a linear elliptic equation,
\begin{align}
  \label{eq:psi-eqs}
  -2\nabla\cdot \left( |K_\e \phi^{n+1}(\bx)|^2 \nabla \psi\right)-\frac{2}{\e}\phi^{n+1}(\bx)\left(1-\phi^{n+1}(\bx)\psi(\bx)\right)+\gamma(\psi-\phi^{n+1}+P^n)=0
\end{align}
with homogeneous Neumann boundary conditions on $\p\Omega$.

\begin{algorithm*}[t]
\floatname{algorithm}{Algorithm}
\caption{Image Inpainting with Model \eqref{eq:opt-elastic-l2}}
\label{alg:inpainting-l2}
\begin{algorithmic}[1]
\STATE Input: given data $\phi_0$ on $\Omega_I$, parameters $\gamma, \e$, initial values $\psi^0, P^0$, error tolerance $\varepsilon$.
\REPEAT
\STATE Update $\phi^{n+1}$ by solving \eqref{eq:phi-l2}.
\STATE Update $\psi^{n+1}$ by solving \eqref{eq:psi-eqs}.
\STATE Update $P^{n+1}$: $P^{n+1}= P^{n}+\psi^{n+1}-\phi^{n+1}.$
\UNTIL $\|P^{n+1}-P^n\|<\varepsilon$.
\STATE Output: $\phi=\phi^{n+1}$.
\end{algorithmic}
\end{algorithm*}

\section{Numerical Experiments}
\label{sec:num}

\begin{figure}[ht]
  \centering
  \begin{tabular}{c}
  \includegraphics[width=0.5\textwidth]{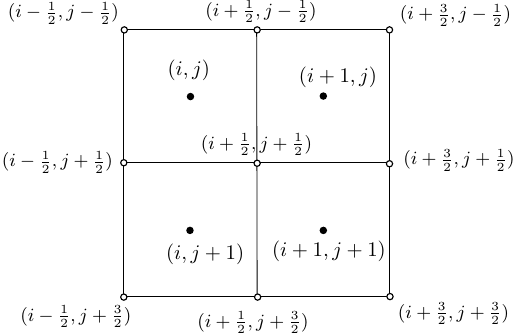}
   \end{tabular}
  \caption{Pixel grid ($\bullet$) and interface grid ($\circ$).}
  \label{fig:grid}
\end{figure}

First, we introduce the discretization of the differential operators. The grid consisting of the centers of the pixels is called the pixel grid ($\bullet$ in Fig. \ref{fig:grid}).
The function $\phi$ is discretized over the pixel grid and
$\phi_{i,j}$ denotes the value of $\phi$ in pixel $(i,j)$. The dual grid of the pixel grid is called the interface grid ($\circ$ in Fig. \ref{fig:grid}).
The gradient $\nabla \phi$ and the convolution $L_\epsilon(\phi)$ are discretized over the interface grid.

More precisely, with $h$ the mesh size, the gradient is discretized as
\begin{align}
  \label{eq:dis-grad-x}
  (\nabla_h^x\phi)_{i+\frac{1}{2},j+\frac{1}{2}}=&\,\frac{1}{2h}\left(\phi_{i+1,j}-\phi_{i,j}+\phi_{i+1,j+1}-\phi_{i,j+1}\right)\\
\label{eq:dis-grad-y}
(\nabla_h^y\phi)_{i+\frac{1}{2},j+\frac{1}{2}}=&\,\frac{1}{2h}\left(\phi_{i+1,j+1}-\phi_{i+1,j}+\phi_{i,j+1}-\phi_{i,j}\right)
\end{align}
Given the discretization of the gradient, the discrete divergence is defined such that $-\nabla_h\cdot$ is the adjoint operator of the discrete gradient,
\begin{align}
  (\nabla_h\cdot D)_{i,j}=&\,\frac{1}{2h}\left(D^x_{i+\frac{1}{2},j-\frac{1}{2}}-D^x_{i-\frac{1}{2},j-\frac{1}{2}}+D^x_{i+\frac{1}{2},j+\frac{1}{2}}-D^x_{i-\frac{1}{2},j+\frac{1}{2}}\right)\nonumber\\
&+\,\frac{1}{2h}\left(D^y_{i+\frac{1}{2},j+\frac{1}{2}}-D^y_{i+\frac{1}{2},j-\frac{1}{2}}+D^y_{i-\frac{1}{2},j+\frac{1}{2}}-D^y_{i-\frac{1}{2},j-\frac{1}{2}}\right).\nonumber
\end{align}

\begin{figure}[ht]
  \centering
  \begin{tabular}{ccccc}
$\alpha=1,\beta=100$&& $\alpha=1,\beta=100$&&$\alpha=1,\beta=80$\\
&&&&\\
  \includegraphics[width=0.2\textwidth]{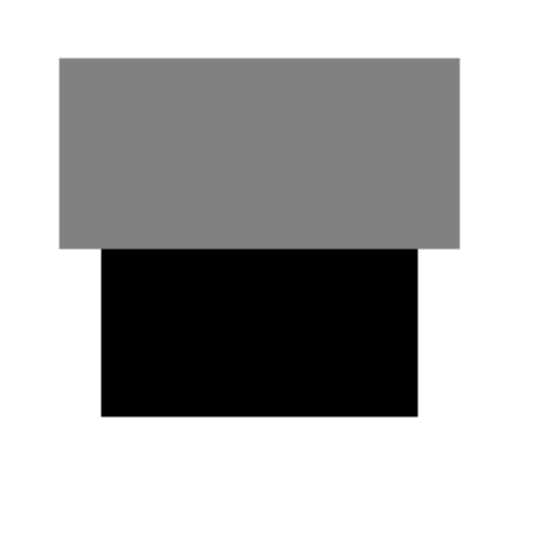}&&
\includegraphics[width=0.2\textwidth]{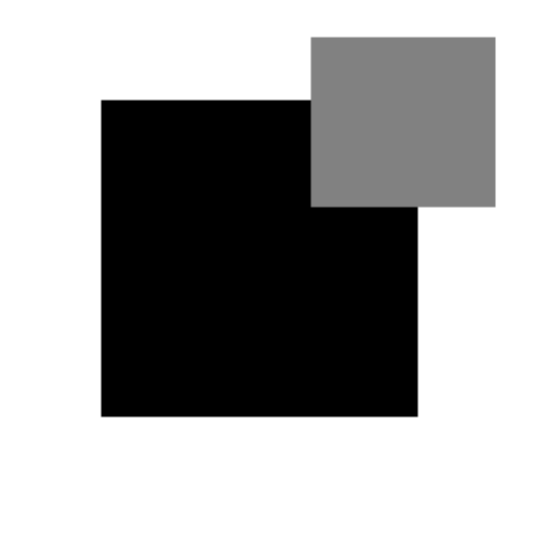}&&
\includegraphics[width=0.2\textwidth]{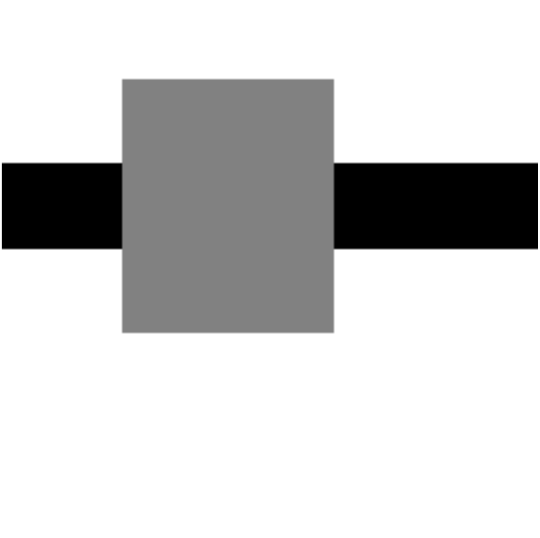}\\
\includegraphics[width=0.2\textwidth]{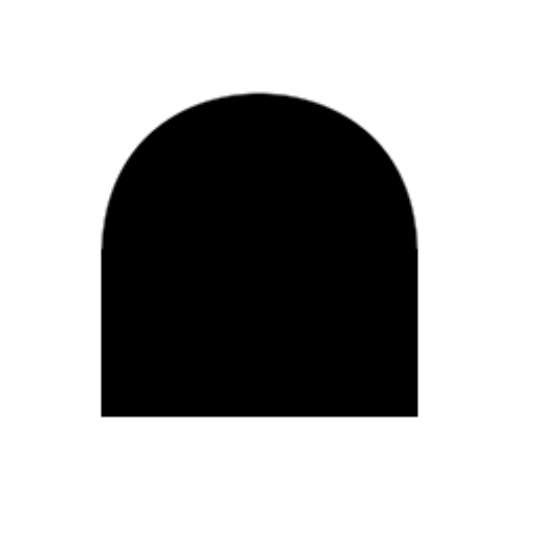}&&
\includegraphics[width=0.2\textwidth]{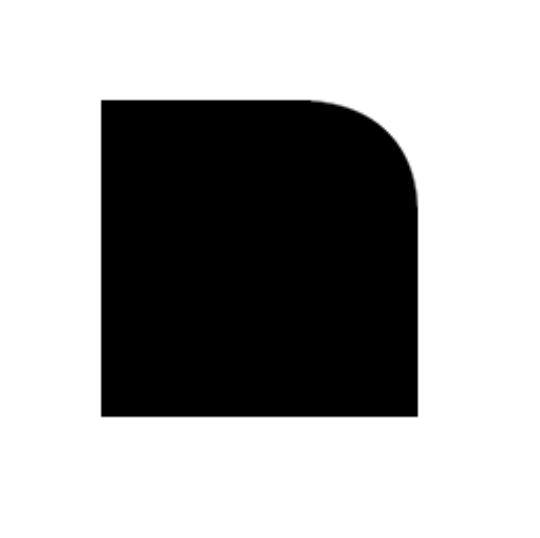}&&
\includegraphics[width=0.2\textwidth]{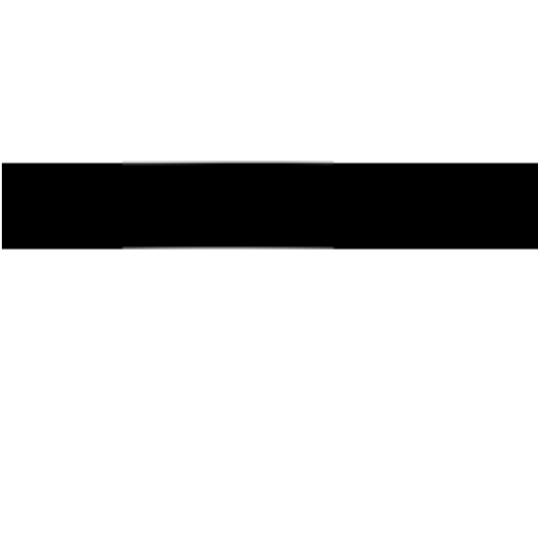}\\
&&&&\\
$\alpha=1,\beta=10$&& $\alpha=1,\beta=10$&&$\alpha=1,\beta=100$\\
&&&&\\
\includegraphics[width=0.2\textwidth]{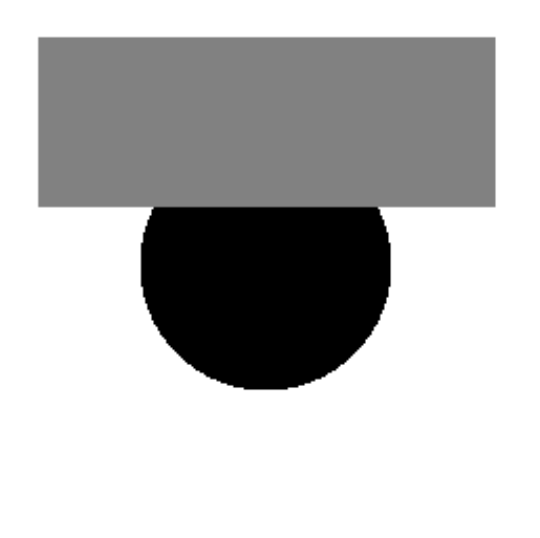}&&
\includegraphics[width=0.2\textwidth]{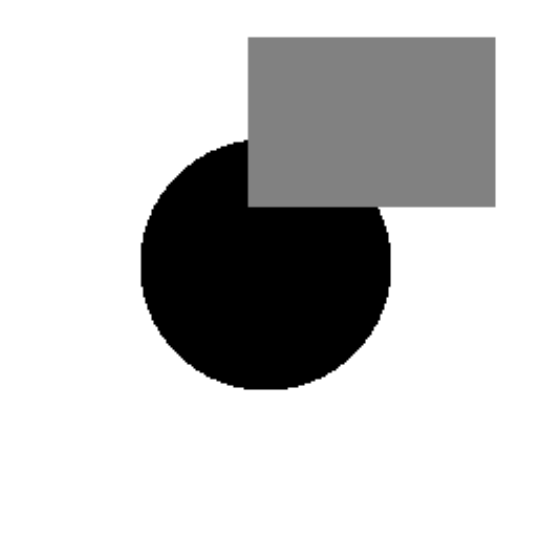}&&
\includegraphics[width=0.2\textwidth]{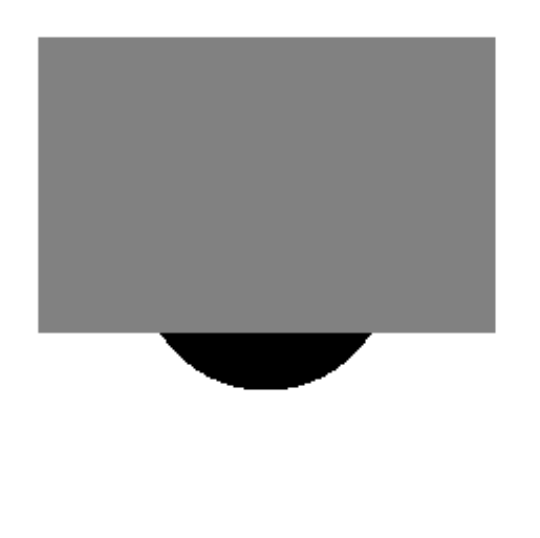}\\
\includegraphics[width=0.2\textwidth]{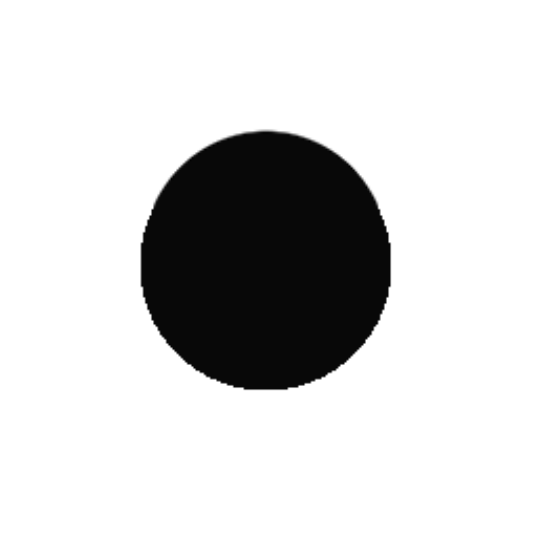}&&
\includegraphics[width=0.2\textwidth]{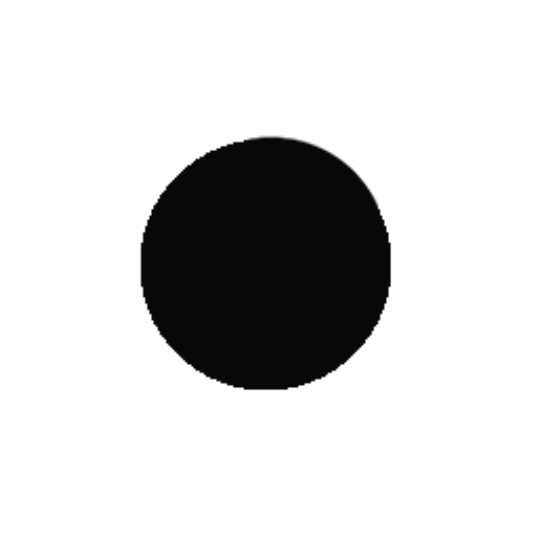}&&
\includegraphics[width=0.2\textwidth]{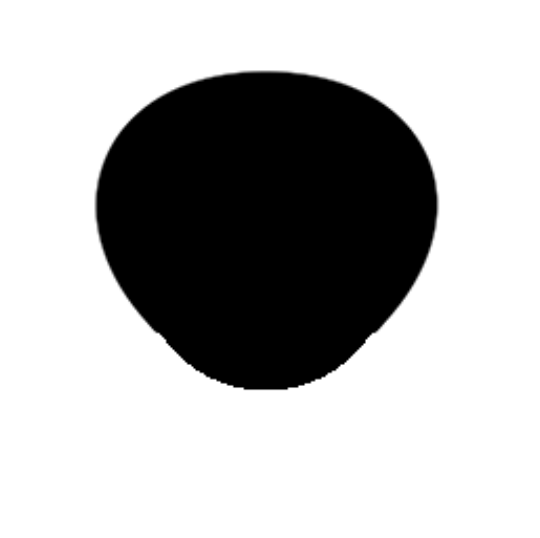}
   \end{tabular}
\vspace{3mm}
  \caption{Inpainting results using Algorithm \ref{alg:inpainting}. The 1st and 3rd rows show the damaged images, in which the missing region
$\Omega\backslash\Omega_I$ is shown in gray, while
the 2nd and 4th rows show the restored images. The parameters $\alpha$ and $\beta$ used for each example are listed above the corresponding column.
}
  \label{fig:resl-inpainting}
\end{figure}

\begin{figure}[ht]
  \centering
  \begin{tabular}{ccccc}
$\gamma=1500$&& $\gamma=2000$&&$\gamma=300$\\
&&&&\\
  \includegraphics[width=0.2\textwidth]{data_box_12.pdf}&&
\includegraphics[width=0.2\textwidth]{data_box_14.pdf}&&
\includegraphics[width=0.2\textwidth]{data_strip.pdf}\\
\includegraphics[width=0.2\textwidth]{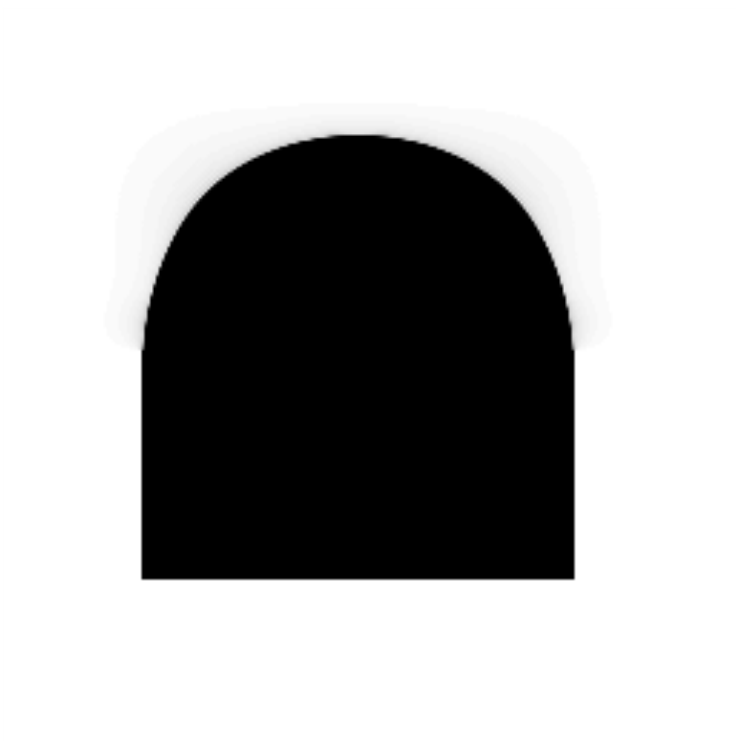}&&
\includegraphics[width=0.2\textwidth]{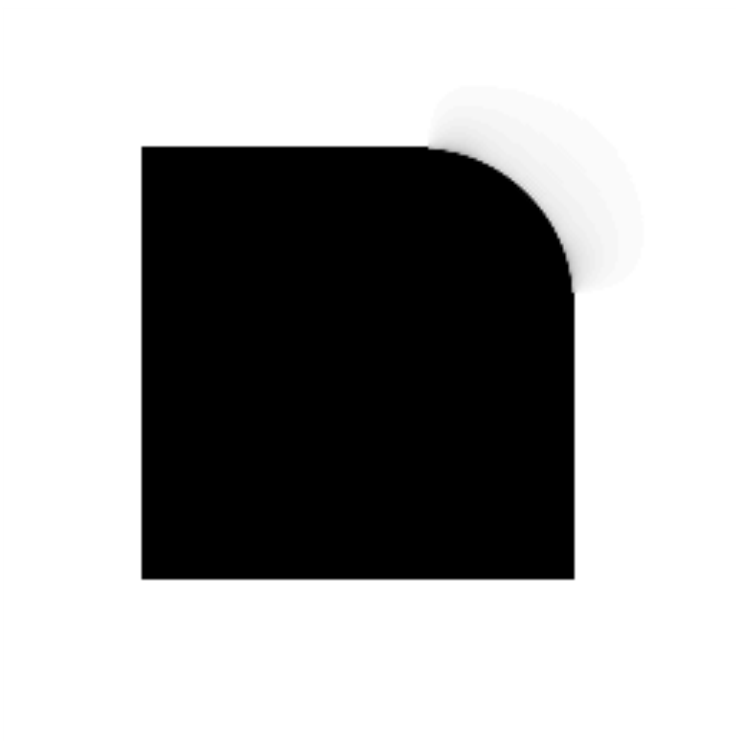}&&
\includegraphics[width=0.2\textwidth]{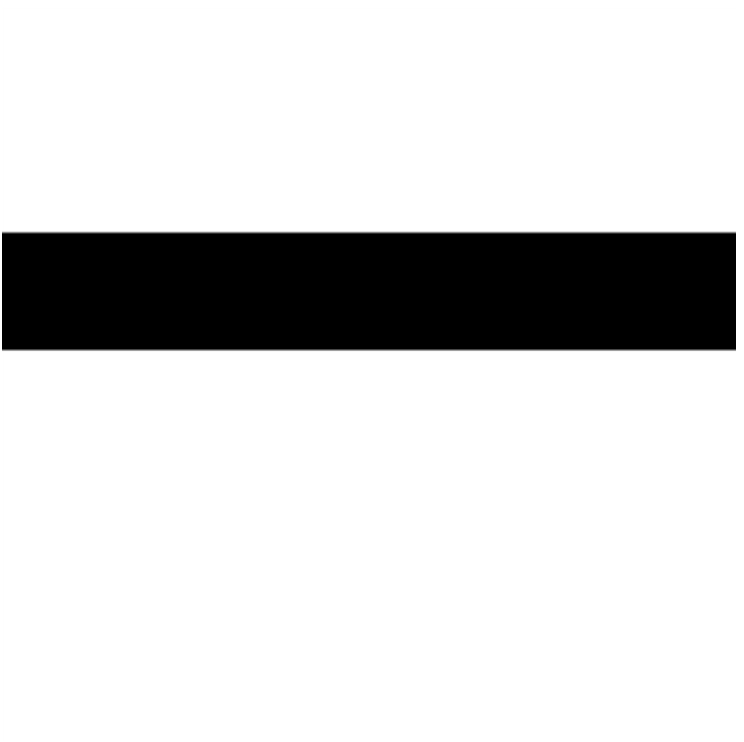}\\
&&&&\\
$\gamma=1000$&& $\gamma=800$&&$\gamma=1200$\\
&&&&\\
\includegraphics[width=0.2\textwidth]{data_disk_12.pdf}&&
\includegraphics[width=0.2\textwidth]{data_disk_14.pdf}&&
\includegraphics[width=0.2\textwidth]{data_disk_34.pdf}\\
\includegraphics[width=0.2\textwidth]{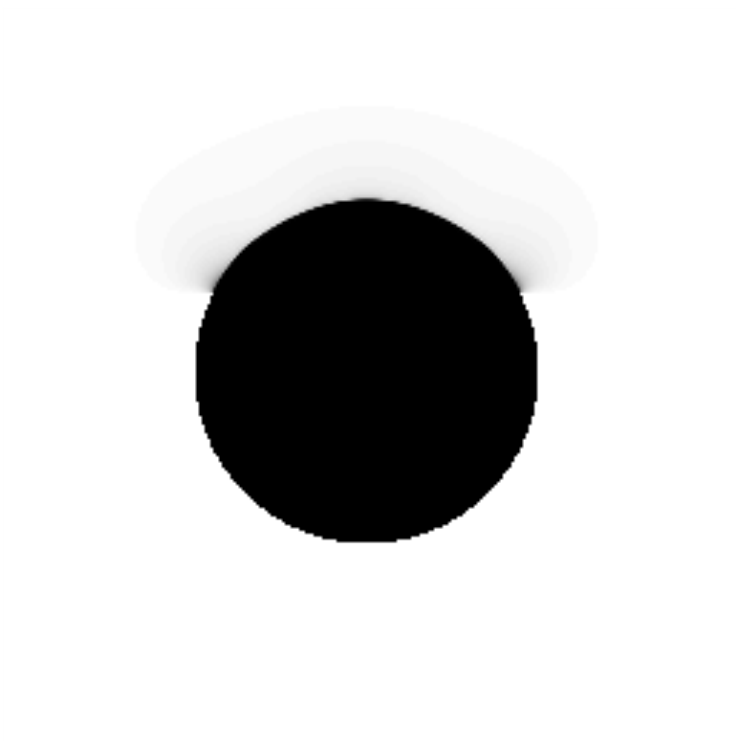}&&
\includegraphics[width=0.2\textwidth]{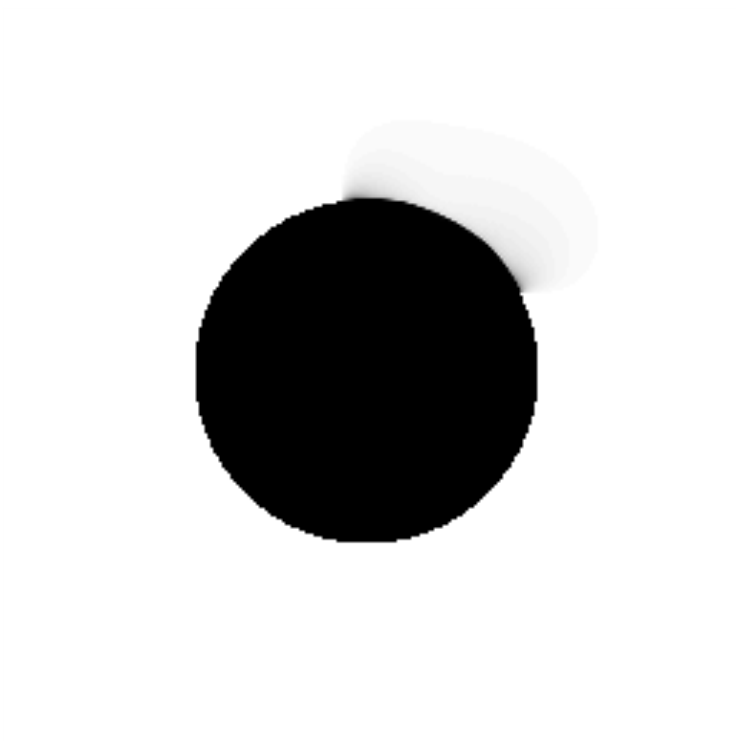}&&
\includegraphics[width=0.2\textwidth]{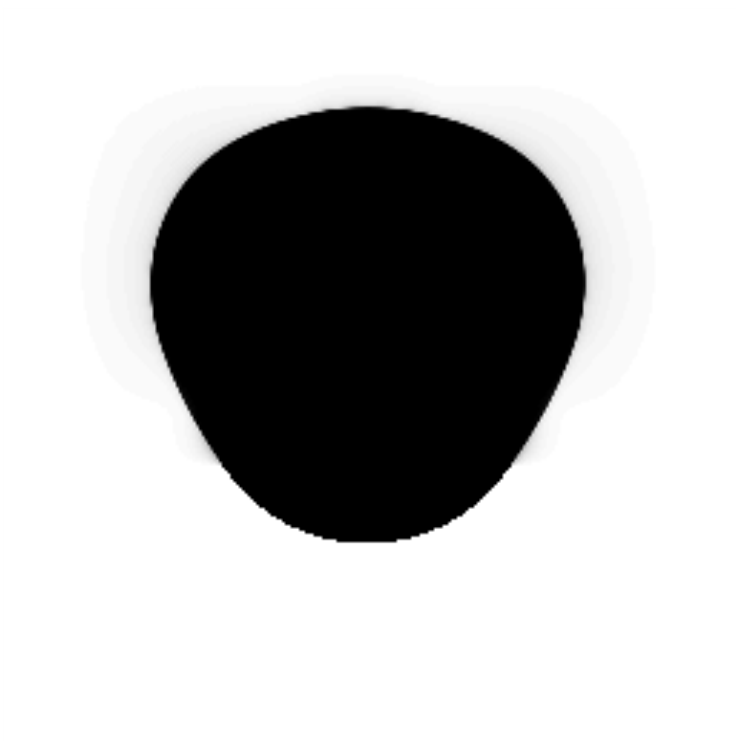}
   \end{tabular}
\vspace{3mm}
  \caption{Inpainting results using Algorithm \ref{alg:inpainting-l2}. The 1st and 3rd rows show the damaged images, in which the missing region
$\Omega\backslash\Omega_I$ is shown in gray, while
the 2nd and 4th rows show the restored images. The parameter $\gamma$ used for each example is listed above the corresponding column.
}
  \label{fig:resl-inpainting-l2}
\end{figure}

In the computations, we use the very simple kernel function \eqref{eq:kernel-R},
\begin{align*}
  R(r)=\left\{\begin{array}{cc}
1,& 0\le r< 1,\\
0,& r\ge 1,
\end{array}
\right.
\end{align*}
for which $D_0=1$ and $C_0=\frac{2}{3\pi}$. The convolutions with $R_\e$ are evaluated by FFT.
The images have resolution $256\times 256$ and the computational domain is normalized to $\Omega=[0,1]^2$, so that $h=1/256$.
The parameter $\e$ is reduced gradually from $0.15$ to $0.06$.
Fig. \ref{fig:resl-inpainting} and \ref{fig:resl-inpainting-l2} show the inpainting results obtained with Algorithm \ref{alg:inpainting} and Algorithm \ref{alg:inpainting-l2}.

\clearpage
\section{Discussion and Limitations}
\label{sec:limit}
In the model proposed in this paper, there are still some issues unresolved. First, the curvature formula \eqref{eq:curvature-intro} is only valid for $\bx\in \Gamma$.
If $\bx$ is off $\Gamma$, a calculation similar to the one in Section \ref{sec:curvature} shows that, in addition to the curvature,
there is another term related to the distance from $\bx$ to $\Gamma$. More precisely,
\begin{equation}
  \label{eq:curvature-off}
  \frac{1}{\e}\int_{\Omega} R_\e(\bx,\mathbf{y})\phi(\mathbf{y})\mathd \mathbf{y}=-C_0\kappa(\bx_0)+O\left(\frac{d(\bx)}{\e^2}\right),
\end{equation}
where $d(\bx)$ is the distance from $\bx$ to the interface $\Gamma$, i.e. $d(\bx)=\inf_{\by\in \Gamma} |\bx-\by|$, and $\bx_0$ is the corresponding closest point on $\Gamma$.
The derivation of the above equation can be found in the Appendix.
In the computation, instead of the binary constraint $\phi=\pm1$, we use the relaxed interval constraint $\phi\in [-1,1]$. With the interval constraint,
the interface is not sharp. It becomes a narrow layer of width $O(\e^2)$. Due to the distance term $\frac{d(\bx)}{\e^2}$, it is not clear whether the proposed model
converges to the elastica model as $\e$ goes to $0$. More theoretical study is required to understand the relation between the proposed model and the elastica model.

Another issue is that the function $\phi$ is not a sharp binary function. It has a transition layer of width $O(\e^2)$. To resolve this layer, we need a very high resolution. Uniformly increasing the resolution
is not practical, while adaptively refining the mesh near the interface seems to be a good choice.

\appendix
\section{Derivation of Eq. \eqref{eq:curvature-off}}
The calculation of \eqref{eq:curvature-off} is similar to the one in Section \ref{sec:curvature}.
For $\bx\notin \Gamma$, let $\bx\in \Omega^+$ and $d=\inf_{\by\in \Gamma}|\bx-\by|>0$. As in Section \ref{sec:curvature}, we abbreviate
$$\hat R_\e(\by)=\bar{R}\left(\frac{|\bx-\by|^2}{\e^2}\right).$$
Since the ball $B_\sigma(\bx)$ is now entirely contained in $\Omega^+$ for
$\sigma<d$, the Gauss formula gives
\begin{align}
  \label{eq:curv-1-d}
&  \pi D_0\e\int_{\Omega} R_\e(\bx,\mathbf{y})\phi(\mathbf{y})\mathd \mathbf{y}
=\lim_{\sigma\rightarrow 0}\pi D_0\e\int_{\Omega\backslash B_\sigma(\bx)} R_\e(\bx,\mathbf{y})\phi(\mathbf{y})\mathd \mathbf{y}\nonumber\\
=& -2\int_{\Gamma} \bn(\by)\cdot\left(\frac{\bx-\by}{\e} \right)\hat R_\e(\by)\,\mathd S_\by
-\lim_{\sigma\rightarrow 0}\frac{2\pi \sigma^2}{\e} \,\bar{R}\left(\frac{\sigma^2}{\e^2}\right).
\end{align}
It is easy to verify from \eqref{eq:R-bar} that
\begin{equation}
  \lim_{\sigma\rightarrow 0}\frac{2\pi \sigma^2}{\e} \,\bar{R}\left(\frac{\sigma^2}{\e^2}\right)=-\pi\e D_0.\nonumber
\end{equation}
In particular, if $d>\e$ then $\Gamma$ lies outside the support of $\bar R(|\bx-\cdot|^2/\e^2)$ and \eqref{eq:curv-1-d} reduces to
$\int_{\Omega} R_\e(\bx,\by)\phi(\by)\mathd \by=1$, as it should be.

Denote
$\bx_0=\arg\inf_{\by\in \Gamma}|\bx-\by|$, so that $\bx-\bx_0=-d\,\bn(\bx_0)$ since $\bx\in\Omega^+$. The integral in \eqref{eq:curv-1-d} can be split into two terms:
\begin{align}
  &-2\int_{\Gamma} \bn(\by)\cdot\left(\frac{\bx-\by}{\e} \right)\hat R_\e(\by)\,\mathd S_\by\nonumber\\
  =&\frac{2d}{\e}\int_{\Gamma} \bn(\by)\cdot \bn(\bx_0) \,\hat R_\e(\by)\,\mathd S_\by
  -2\int_{\Gamma} \bn(\by)\cdot\left(\frac{\bx_0-\by}{\e}\right)\hat R_\e(\by)\,\mathd S_\by.
\end{align}
The second term can be computed using the same method as in Section \ref{sec:curvature} and gives $-\pi D_0\e^2C_0\kappa(\bx_0)+O(\e^3)$. The first term gives the
contribution proportional to $d/\e^2$. Writing $\bn(\by)\cdot\bn(\bx_0)=1+O(s^2)$, where $s$ is the arclength of $\Gamma$ measured from $\bx_0$,
\begin{align}
  \frac{2d}{\e}\int_{\Gamma} \bn(\by)\cdot \bn(\bx_0)\,\hat R_\e(\by)\,\mathd S_\by
  =& \frac{2d}{\e}\int_{\Gamma} \hat R_\e(\by)\,\mathd S_\by+\frac{2d}{\e}\int_{\Gamma} O(s^2)\, \hat R_\e(\by)\,\mathd S_\by.
\end{align}
Using a simple scaling argument, the second term is $O(d\e^2)$, i.e. it contributes $O(d)$ to \eqref{eq:curvature-off}.
To simplify the calculation of the first term, we use the simple kernel function \eqref{eq:kernel-R}, for which $D_0=1$ and $\bar R(r)=\frac{1}{2}(1-\frac{1}{r})$ for
$0\le r<1$. Approximating $\Gamma$ near $\bx_0$ by its tangent line, so that $|\bx-\by|^2\approx d^2+s^2$, we obtain
\begin{align*}
  \frac{2d}{\e}\int_{\Gamma} \hat R_\e(\by)\,\mathd S_\by\approx\frac{2d}{\e}\int_{0}^{\sqrt{\e^2-d^2}} \left(1-\frac{\e^2}{d^2+s^2}\right)\mathd s
  =2\left(d\sqrt{1-\frac{d^2}{\e^2}}-\e \arctan\sqrt{\frac{\e^2}{d^2}-1}\right).
\end{align*}
Putting all the estimates together and dividing by $\pi D_0\e^2=\pi\e^2$, we get
\begin{align}
  \frac{1}{\e}\int_{\Omega} R_\e(\bx,\mathbf{y})\phi(\mathbf{y})\mathd \mathbf{y}
  =&-C_0\kappa(\bx_0)+\frac{1}{\e}+O(d)+O(\e)\nonumber\\
  &+\frac{2}{\pi\e^2}\left(d\sqrt{1-\frac{d^2}{\e^2}}-\e \arctan\sqrt{\frac{\e^2}{d^2}-1}\right).\nonumber
\end{align}
Expanding the middle term for $d\ll\e$ and using $\arctan\sqrt{\e^2/d^2-1}=\arccos(d/\e)$, the two terms $\frac{1}{\e}$ cancel and we are left with
\begin{align}
  \frac{1}{\e}\int_{\Omega} R_\e(\bx,\mathbf{y})\phi(\mathbf{y})\mathd \mathbf{y}
  =-C_0\kappa(\bx_0)+\frac{4d}{\pi\e^2}-\frac{2d^3}{3\pi\e^4}+O(d)+O(\e),\nonumber
\end{align}
which is \eqref{eq:curvature-off}. For $\bx\in\Omega^-$ the same computation applies with the sign of the distance term reversed.

\bibliographystyle{abbrv}
\bibliography{reference}

\end{document}